\documentclass[letterpaper]{article}

\usepackage[utf8]{inputenc}
\usepackage[a4paper,bindingoffset=0.2in,%
            left=1in,right=1in,top=1in,bottom=1in,%
            footskip=.25in]{geometry}
\usepackage[colorlinks = true,
            linkcolor = blue,
            urlcolor  = blue,
            citecolor = blue,
            anchorcolor = blue]{hyperref}
\usepackage{graphicx}
\usepackage{pdfpages}
\usepackage[numbers,sort,compress]{natbib}
\usepackage{graphicx}

\title{Glamour muscles: why having a body is not what it means to be embodied}
\author{Shawn L. Beaulieu$^{1}$ and Sam Kriegman$^{2}$ \\
\mbox{}\\
$^1$University of Vermont, Burlington, VT 05446 \\
$^2$Northwestern University, Evanston, IL 60208 \\
\href{mailto:sle.beaulieu@gmail.com}{\color{blue}sle.beaulieu@gmail.com}
} 

\begin{document}
\maketitle

\begin{abstract}

Embodiment has recently enjoyed renewed consideration as a means to amplify the faculties of smart machines. Proponents of embodiment seem to imply that optimizing for movement in physical space promotes something more than the acquisition of niche capabilities for solving problems in physical space. However, there is nothing in principle which should so distinguish the problem of action selection in physical space from the problem of action selection in more abstract spaces, like that of language. Rather, what makes embodiment persuasive as a means toward higher intelligence is that it promises to capture, but does not actually realize, contingent facts about certain bodies (living intelligence) and the patterns of activity associated with them. These include an active resistance to annihilation and revisable constraints on the processes that make the world intelligible. To be theoretically or practically useful beyond the creation of niche tools, we argue that ``embodiment'' cannot be the trivial fact of a body, nor its movement through space, but the perpetual negotiation of the function, design, and integrity of that body—that is, to participate in what it means to \textit{constitute} a given body. It follows that computer programs which are strictly incapable of traversing physical space might, under the right conditions, be more embodied than a walking, talking robot. 

\end{abstract}

\section*{Meet the new paradigm. Same as the old paradigm.}

The accomplishments of artificial intelligence are legion. But so too are its deficiencies. To say this is not to impugn the hard work of those who advanced the field to its present state, nor to doubt the material gains this work has secured. It is instead to acknowledge that every technology has costs and benefits that are intrinsic to both its design and the context in which it functions \cite{Ong1992}. While we have witnessed tremendous progress in the generation of text and images \cite{Brown2020, Chowdhery2022}, event classification and forecasting \cite{Lecun1998, ResNet, Ravuri2021}, planning \cite{Silver2018, Jaderberg2019, Perolat2022, Bakhtin2022}, product design \cite{Mirhoseini2021, Li2022}, and scientific discovery \cite{Jumper2021, Chen2022, Champion2019, Assael2022, Fawzi2022} we have also seen that AI programs behave in ways that cast doubt on their reliability \cite{Athalye2018, Kurakin2016, Marcus2020, Dantella2023, Dziri2023}, and show a persistent brittleness that curbs their widespread utility \cite{french1999, Kirkpatrick2017, Masse2018, Xu2019}. Researchers hoping to address these issues often look to living intelligence for inspiration, and among the most distinguishing features of living intelligence is the collection of traits referred to as \textit{embodiment}.

Embodied machines are meant to be systems whose expanded component function exhibits or obtains special qualities that are inaccessible to more constrained systems \cite{Duan2021, Driess2023, EmbodiedWorkshop2022, SayCan2022, Xiang2023, Pfeifer2007, Kober2008, Man2019, Glenberg1997, Anderson2003, Dreyfus92}. It is thought that embodied machines will better mirror the traits of their living counterparts insofar as classically cognitive phenomena like memory, perception, and representation are causally wedded to bodily action. Autonomous movement among and between otherwise immobile parts is what enables these systems to acquire novel information about the world they inhabit, and to change its state by way of interaction. Observations are thus not merely subject to accurate or useful processing; rather, the details of observation are actively generated and transformed by bodily action. Embodiment is then the total imbrication of action and perception, and is thought to bestow upon machines the ability to grasp the \textit{implications} of what they observe and the \textit{consequences} of what they do.

To speak in these terms, however, is to frame the prospective benefits of embodiment as relating either to the type of action an embodied machine performs (e.g. mechanical rotation versus image classification), or to how this action affects the distribution of inputs the machine receives. That different actions have different effects, and therefore different uses, is not a special contribution of embodiment \textit{per se}, but a generic fact about technology in general. Thus, to argue that embodied machines afford us the capacity to automate physical processes \cite{Duan2021, Driess2023} is to say nothing more than that there are circumstances in which we would do well to use an image classifier, a language model, or a robot. Meanwhile, to argue that what embodied machines do, and what disembodied machines cannot do, is control the distribution of inputs they receive \cite{Duan2021, EmbodiedWorkshop2022, Kagan2022, SayCan2022} is only to make an argument about the possible efficiency of the embodied approach. If learning what is entailed by certain events is still a function of inputs which are specified in advance, even if these are conditioned on the actions a machine performs, then we have not defined a qualitatively different methodology. All that disembodied programs need to acquire such mechanistic knowledge is a sufficiently large dataset of relevant observables from which to infer cause and effect. Great effort might be necessary to produce such a dataset, and embodied machines might thus more rapidly infer causal relations from the actions they elect to perform \cite{Bongard2011, Xiang2023}, but the measure of their utility still consists in how well they map a given set of inputs to a given set of outputs.

Since the data on which embodied machines are trained need not be supplied prior to their deployment, they appear to us as open-ended, or indeterminate, in a way that disembodied machines do not. What this obscures is that the specification of relevant inputs, and the definition of permissible actions, is encoded in the very design of such a machine, which is not significantly altered by the activity the machine itself realizes. One might object that some machines are programmed with the ability to add, remove, and otherwise modulate system components \cite{seo2019modular, Ha2022}, and so appear to escape this characterization, but decisions regarding which phenomena to use or model, and how to do so, are clearly delineated by the engineering which defines them as the machines they are%
\footnote{It is worth quoting the late cyberneticist, Gordon Pask, at length on this issue: ``A valve, for example, accepts only an electrical input and provides an amplified electrical output. If it also responds to temperature or vibration, it is to this extent a bad valve. The logical simplicity of the computer model is a consequence of being able to put one's finger upon a component which performs a known function and to reject the imperfections as irrelevant...When trying to construct a physical model of a [self-organizing system] we are beset with a peculiar difficulty...The logical requirements force us to use media such that, when a physical model is constructed, we cannot specify components which have a well defined function, and we cannot separate inputs and outputs into a set which are relevant and a set which may be discounted. It is inherent in the logical character of the [self-organizing system] that all available methods of organization are used, and that it cannot be realized in a single reference frame. Thus, any of the tricks which the physical model can perform, such as learning and remembering, may be performed by one or all of a variety of mechanisms, chemical or electrical or mechanical. Thus, however much we try, we cannot achieve an electrical model or a mechanical model or a chemical model of a self-organizing system. Any physical model necessarily includes them all in varying degrees.'' \cite{Pask1960}}
They therefore depend on some external process by which relevant factors are specified \cite{Pask1960, Cariani2007, Cariani1989}. Although the \textit{shape} of the input distribution is not given to embodied machines in advance, but results from their interaction with the environment, \textit{what phenomena} populate the distribution is rigidly constrained by what the machine is made to perceive and how it is made to act. This may or may not result in efficiency gains relative to a disembodied machine with access to more or better data, but it will yield real benefits insofar as we need a machine that moves. But nor does defining embodiment as the push and pull of action and perception necessarily imply a relation to physical movement. Provided that language models directly participate in the social process of determining how language is used, rather than merely how to use language given its social determination, they achieve a linguistic analog to physical embodiment \cite{Santoro2021}. Physically embodied machines are thus distinct only in what aspects of the world they are licensed to observe and alter. And just as the manipulation of physical objects might ground \cite{Barsalou2008} or supplement \cite{Ramesh2021} certain cognitive faculties, so too might the use and manipulation of language augment visual comprehension in image models. There is no a priori reason to believe that physical movement as such is uniquely informative, or causally efficacious, relative to other kinds of action, except insofar as it's necessary to solve certain problems. 

Far from transcending orthodoxy, such a shallow notion of embodiment merely repackages the idea of intelligence as reinforced selection \cite{Pask1958}. But this is not altogether undesirable. Skeptics of embodiment as an alternative paradigm can rest easy knowing that nothing new or supernatural is being proposed, while proponents of embodiment can celebrate the fact that, under certain conditions, their methods might secure valuable gains. To leave it at that, however, would be to forfeit the promise of a genuine alternative to existing theories and methods, and would reduce physically embodied machines to niche tools for solving or reasoning about problems in physical space. Yet we know that the capacities of living intelligence aren't exhausted by mere mobility. What other principles can we draw on to make machines that are meaningfully different from those presently in existence? And what reasons do we have to pursue such a project?

\section*{Taking embodiment seriously.}

\begin{figure*}[!t]
\centering
\includegraphics[width=1\linewidth]{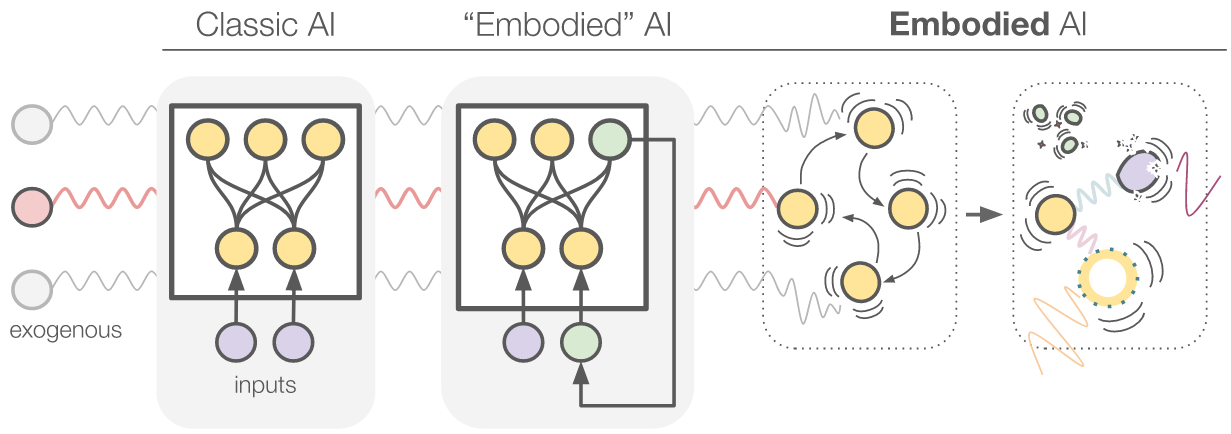}
\caption{Both classic AI (image classifiers, language models, etc.) and ``embodied'' AI (robots) map a given set of inputs to a given set of outputs according to a specific set of rules. Neither the rules, nor what counts as an input or output, are modified by either type of machine unless an additional set of rules permits relevant modifications. But for \textbf{embodied} AI, what counts as an input or output, and what operations it is capable of realizing, would not be conclusively specified prior to its deployment. These would instead emerge through its interaction with arbitrary features of its environment, from which the system itself cannot be cleanly separated.}
\label{figure_0}
\end{figure*}

A constitutive property of life is that in forming and maintaining itself it can, within the limits of its powers, construct or exploit whatever affords it the means to do so \cite{Oyama2000book, Froese2009, KauffmanOrigins1993, Pask1958, Pask1960, Pask1961, Varela1974, Cariani2007, Nicholson2019, EmmonsBell2019, Ashby1960, Layzell2001, Bird2003}. If it is adaptive for an organism to develop an affinity for sound, or attend to fluctuations in ambient temperature, then all that stands in its way are the viability of the mechanism involved and the likelihood of its discovery. What passes for embodiment in machine learning ignores the details of how bodies are formed, adapted, and maintained \cite{Duan2021, Driess2023, EmbodiedWorkshop2022}. This labor is instead performed by humans, who establish the conditions for its comparatively simple methods to work. We propose to call these machines \textit{glamour muscles}, because in their effort to emulate life they merely copy or refine its superficial form and function; they don't meaningfully participate in their own design, function, and maintenance. Herein lies the conceptual slippage between ``embodiment'' as a description of living intelligence, and ``embodiment'' as the realization of physical movement in otherwise immobile and inert machines. Natural systems we would regard as paradigmatically embodied don't just have bodies inside of which cognition occurs, nor is their physiology conscripted to perform an immutable, or deliberately bounded, set of operations for satisfying pre-given ends \cite{Varela1974, Pharoah2020, Logan2012}. Rather, the definition of a body—and all that it can sense, compute, and perform—arises from a perpetual, and historically specific, negotiation with the environment; which itself undergoes continual change and redefinition \cite{Cariani1989, Cariani2001,  Oyama2000book, LevinsLewontin1987}. This is not to say that organisms are infinitely plastic, or that we should aspire to build machines that are free from constraint. To the contrary, some manner of constraint above and beyond what is physically permitted provides the necessary conditions for life's possibility \cite{Hagglund2019book, Cariani1989, Cariani2001, Pattee2001}. Rather, the essential question is whether the space of possible actions and percepts for a given system can be plausibly inferred from how it presently works \cite{Pask1961, Roli2022, Cariani1989, Rosen1986}. For existing AI the answer is yes, because the logic by which it operates is fully determined by constraints it cannot challenge or transform \cite{Pask1960, Cariani2007, Roli2022}. No existing language model will ever, without expert modification, commit itself to anything other than language, even if it would be adaptive to do so. Precisely why a language model produces a given string may be opaque to us, but that it produces strings (to any degree of coherence), and not salt water taffy, is a consequence of it being designed for language use.

Yet satisfying this condition of unpredictability in practice does not require unpredictability in principle, or the possession of an immaterial will. It depends on a fundamental \textit{indeterminacy}%
\footnote{We might here distinguish between epistemic uncertainty, which reflects the inadequacy of whatever description we use to characterize a given system, and practical indeterminacy, which is a \textit{source} of epistemic uncertainty. Epistemic uncertainty may never be completely resolved even for practically \textit{determinate} systems, owing to the provisional nature of all scientific hypotheses \cite{PopperBook}. But this problem is only compounded by practically \textit{indeterminate} systems, because it cannot be entirely known what features of the world, including what aspects of their own dynamics, these systems are, or will be, relevantly coupled to \cite{Pask1961, Layzell2001, Bird2003, Levin2022}. Predicting or explaining the dynamics of structurally determinate systems can be done more or less suitably with comparably determinate models, but indeterminate systems require descriptions that change with those structural alterations that redefine the system under investigation \cite{PaskFoerster1960, Pask1960, Rosen1986}.}
of form and function,
which resolves itself through material interaction—both with the surrounding environment, to the extent we wish to make such a distinction, and among its ever shifting and mutable parts \cite{Wood1995book, Oyama2000book, LevinsLewontin1987, Cariani1989, Froese2019, Pask1958}.
Even benign mechanical operations for living things involve the risk of annihilation that must be mitigated by the very mechanisms that compose these operations, and can only be made sense of under conditions these operations help to construct and maintain. Whereas the most elementary operations for machines are only made possible or intelligible by the prior commitment of human labor. Given this labor, the social and material conditions that enable smart machines are causally isolated from the activity these machines themselves realize. When we say, for instance, that a neural network takes pictures as input and produces class labels as output, we are enforcing a special set of constraints on a system that define what it is (a classifier) and what it does (classify pictures). The process of classifying pictures, which is the only activity the system is allowed to perform, has no power to directly alter these constraints. Although we, observing its performance and finding it wanting, may choose to do so. What then defines the system, and what enables the activity it realizes, is independent of activity internal to the system (the act of classifying pictures). This is true even in those cases where an algorithm has been programmed to alter certain of its constraints, because what the algorithm is permitted to alter is not itself open to modification \cite{Schmidhuber1993, Layzell2001, Mueller2015, AB1, Bongard2011, Lipson2000, Miller2002, Mirhoseini2021, Li2022, Ha2022}. And since the knowledge and resources required to make such specifications are not uniformly distributed in the population, it is not just our technology but \textit{we} who are captive to the design choices of those who build it. Breaking the distinction between the specification of constraints and the activity these constraints enable will confer upon \textit{machines} the capacity to develop fundamentally new traits, and upon everyday \textit{users} the capacity to shape the development of these traits by interacting with them. This may help us solve problems for which plausible hypotheses, relevant observables, or appropriate actions are unknown, or for which the problems themselves elude definition \cite{Cariani1989}, while at the same time linking the design and purpose of such machines to the immediate interests of users \cite{Pask1958Namur}.

For this to be possible, the constraints that specify what a given machine is or does must be amenable to alteration by \textit{arbitrary} forces acting on \textit{arbitrary} spatial and temporal scales, which carries the unending risk that such alterations might render the machine inoperable. But the risk of annihilation is both the cost paid by relaxing those constraints on material interaction that usefully limit behavioral variety in machines, and the condition of possibility for the acquisition of novel traits. Without the means to expand the variety of possible states, the variety of possible behaviors is entailed only by the present design, but increased variety means increased risk of error \cite{Pask1961, Pask1961book}. Mere susceptibility to dissolution, however, is insufficient. Whirlpools, souffl\'es, and vulcanized rubber are all things that deteriorate, but are not thereby instances of living intelligence. For the possibility of annihilation to mean anything, for it to entail certain behaviors, there must be \textit{reasons} in light of which one acts to forestall the disintegration of that which one values. It is not a feature of how shaving cream or chatbots are organized that their organization is adapted or maintained to ensure their survival. If we were to withhold the energy needed to power a computer except on occasion it did something we desire, and reward it with energy for every good deed it performed, we would not have trained it to do anything, because acting to obtain energy is not built into the organization of computers. Your phone does not ``care'' if you fail to charge its battery. In a shallow sense, you can program it to care, and make it emit increasingly urgent signals reminding you to plug it in as its battery depletes. You can even make the content of these signals adaptive in some way, in order to increase the probability of charging. But there are no practical consequences for the device itself if it does not receive a charge. Even if it ``dies'' you can plug it in tomorrow and it will work just fine. This is because it is designed not to have any such practical consequences resulting from access to energy, just like it is designed not to have any practical consequences resulting from the texture or tensility of the surface on which it rests. This makes it highly reliable and efficient for certain applications. But because there are no practical consequences resulting from these aspects of the world, they cannot be incorporated into the activity of the device unless doing so is already a feature of its design. Except for the phenomena which are specified as relevant to their function, such devices are characterized by a rigid separation between the world they inhabit and the activity they realize.  

This should not be confused with the idea that machines are state-determined, while living intelligence is not \cite{Froese2019}, but that changes to the state of living intelligence by phenomena they lack the ability to detect or act upon can result in the means to do so if such a change in state affects viability and can be relevantly incorporated into future activity \cite{Cariani1989, Oyama2000book}. To assume that prior knowledge of the anatomical structure which detects light, and how to build it, is necessary for light-sensitivity to emerge is to attribute the logistics of machine design to that of living intelligence. But the capacity to measure and control specific phenomena is a capacity with a specific history—one that cannot be explained by assuming the capacity already exists, and simply must be expressed \cite{Wood1995book}. 
Prior work \cite{Pask1958, Pask1960} 
has shown that if an existing component is not deliberately made to be insensitive to an environmental factor (e.g. light, temperature, vibration, etc.) and if the environmental factor can thus affect its physical state, this component can \textit{develop} sensitivity if the energy supplied for its continued growth and maintenance is conditional on its change of state being useful for detecting or acting upon the environmental factor.%
\footnote{Working with metal circuits grown by electrochemical deposition—a process controlled at a given point in solution by one among many automata competing to maximize the means of circuit growth and persistence—Gordon Pask used this idea to generate structures that acquired, but were not designed to have, an affinity for sonic vibration \cite{Pask1958, Pask1960}. By supplying energy only to those competing automata associated with circuits whose output evidenced the desired sensitivity, automata were compelled by selection to produce circuits that improved on, or amplified, the sensitivity of their peers. Owing to the physical dependence of circuits on their detection of relevant stimuli, and the pressure among automata to secure limited resources, specialized sensors were grown and maintained using components with no pre-given function.}
On the other hand, sensitivity to temperature, unless it is specified in advance, will not emerge in an image classifier, or the hardware on which it runs, even if it could improve its ability to classify images. In short, the developmental history of living intelligence is not devised and nurtured from without, but the historical achievement of its antecedent forms acting in a world alive to their own making. 

Vulnerability in robots \cite{Man2019, Kano2017}, disintegrating architectures \cite{gilpin2008miche}, dynamic configuration of parts \cite{Slavkov2018, li2019particle, Ha2022}, and biobots composed entirely of living tissues \cite{kriegman2020xenobots} have all been reported; but, in general, they only passively contend with the forces of their environment. They don't actively resist the tendency to disorder, or secure conditions for their continued operation. Organisms resist the tides of entropy by growing and maintaining structures appropriate to their circumstances, or else by evoking circumstances appropriate to their capacity for growth and maintenance. Robots cannot yet grow new structures, but they have been made to deform \cite{Man2019} or re-adhere \cite{bai2022autonomous} their remnant structure to rescue behavior lost to injury. Recovering function erased by mechanical injury through exaptation of viable parts has a rich history in robotics \cite{Lipson2000, Bongard2006, kriegman2019automated}. But however much exaptation might mirror the compensatory strategies seen in mature organisms, the development of new and contextually determined structures, as seen in ontogeny, has no real analog in machines. To the extent that artificial constructs exhibit this capacity it is often by a strict division of labor, with chemical and biological processes inserted into traditional machine pipelines—as in the case where novel machine capacities are built by human labor, or when events are detected and redressed by simple lifeforms \cite{DeMarse2005, Sinapayen2017, Masumori2019, Kagan2022, Zhu2018, Kamm2018}. This later category of methods recognizes that survival-seeking systems are inveterate problem-solvers \cite{Pask1961, Pickering2008, Bongard2021, Levin2022, Davies2023}, and that profitable collaboration involves translating problems into a common language, along with the enforcement of conditions that transform activities normally undertaken for the sake of the survival-seeking system into those that satisfy externally imposed ends.%
\footnote{If administered to entities that warrant ethical consideration, we might think this method unpalatable, but it isn't necessary that access to the means of persistence is limited to those occasions in which an exogenous objective is fulfilled. An objective may be endogenously generated, and its fulfillment may be predicated on the grounds of preserving an identity or commitment, rather than physiological integrity. Or an objective may be exogenous only in the sense of not being reducible to egoistic satisfaction, but instead aimed at securing the conditions for all to flourish \cite{Hagglund2019book}. Labor performed for the satisfaction of the exogeneous objective—which answers to, but is not exhausted by, individual aspirations—would then be equivalent to the labor necessary for the realization of one's own freedom. What \textit{is} critical to this method, however, is the possibility of, and resistance to, dissolution—a dynamic which, in machines, is scrupulously avoided, or framed as an obstacle to be overcome \cite{Widrow1962, Kirkpatrick2017}. This is done by sharply constraining their material properties and the scope of their permitted activity.}
Rather than plan and manage every detail of optimization, which requires extensive knowledge and foresight, the inborn competencies of survival-seeking systems can be exploited for highly specific purposes if it is a condition of their existence that these purposes are realized \cite{Pask1958, Pezzulo2016, Bartlett2022}. By analogy, it is precisely that to provide value for an employer is what it means to provide subsistence for yourself that control over the product of your labor can be so precisely wielded \cite{Wood1995book}. Once having constructed machines which can appropriately be called ``embodied'' in the sense we've articulated, this mode of interaction may help structure their behavior since, by virtue of being so embodied, they would no longer be subject to detailed programming. Likewise, employers need not have intimate knowledge about how you come to perform the labor you do, or for what reasons, in order to compel specific acts of labor. Rather, your labor is entailed by the need to secure what is required to live and pursue your goals, owing to the way society has been organized. Different forms of behavior would be entailed by different forms of social organization, and this is no less true for actually embodied machines. 

Thus, if resistance to annihilation is the engine of living intelligence, it might also be used to give material force to social desire. Democratic use and design of technology often consists in collating various user interests into one among many constraints on the decisions of those who build and own technology \cite{Brabham2008, Wagy2015, Mahoor2017}. This tradition is admirable for its commitment to inclusivity at the point of production, but nevertheless falls short of enacting the principle articulated by Gordon Pask that would explode their function as instruments of alienation \cite{Pask1958Namur}. 

In delegating decision-making to the labor of another process (machines, committees, political representatives, etc.) it should be the case that the power and scope of the delegate is commensurate with the practical consequences resulting from the decisions they make. If the practical consequences of delegate decisions are (or believed to be) undesirable, then the function of the delegate should be progressively dissolved, until which time it can prove its decisions will be desirable. Representative action, insofar as it's necessary, then becomes materially possible only if the represented parties stand to be satisfied. Otherwise, both the claim to represent these parties, and the capacity to act on their behalf, are immediately rescinded without the need for deliberation. To represent another party is thus precisely what it means to carry out their interests, rather than to transform those interests into actions consistent with more rational ends, or more congenial to the interests of the representative. 

This has implications for even local problem areas, like AI safety. Concerns around AI alignment \cite{Gabriel2020} often assume that the production and deployment of future technologies will be mediated by existing forms of social organization, with products developed in isolation from the immediate interests of users, only later to be foisted upon them. Without making light of the task before us, it is nonetheless worth considering whether alignment might be better achieved by popular involvement at the point of production, and subsequently maintained by the obligate satisfaction of user desire. Social needs would thereby become arbitrary material forces that act directly on machines, much like light or vibration, rather than something which affects machines only through a given technical interface, political procedure, or indirect market exchange. Modeled on the approach taken by Pask and others, we might then collectively build and liquefy various delegations, with powers of action and discernment that far exceed those of any individual, on the insuperable condition they yield material benefits we all participate in defining. This will involve the construction of actually embodied machines: ones that do not simply traverse physical space, but that shape, and are shaped by, arbitrary material events, including those social factors without which they wouldn't even be intelligible as the machines we take them to be.

\section*{Conclusion}

Embodied machines strike us as being indeterminate in a way that supposedly \textit{disembodied} machines do not, because the distribution of inputs they receive is not specified prior to their deployment, but results from their interaction with the environment. However, what counts as an input or output for embodied machines, as well as the rules for processing them, is conclusively specified by their design. For example, a machine equipped with light sensors will not transform them into smoke detectors or pH meters, nor will a language model elect not to minimize prediction error (unless these are already features of their design). On the other hand, living intelligence has the capacity to determine, by virtue of its own activity rather than by external design, what features of the world—including what features of its own activity—are relevant to the form of life it sustains. It does this by growing and maintaining new structures, and by being open to modification by the world it inhabits, which both increases its risk of error and establishes the conditions for generating novelty. Building machines that approach living intelligence in these respects will result in more flexible devices, whose form and function may be determined by their interaction with users.

\section*{Acknowledgements and Author Statements}

SLB would like to thank Kate Nolfi for discussion and editing, and Kyrill Potapov for reading an early draft. 

\footnotesize
\bibliographystyle{apalike}
\bibliography{bibliography} 

\end{document}